\documentclass[conference]{IEEEtran}
\ifCLASSINFOpdf
\else
\fi
 \usepackage{url}
\usepackage{chngpage}
\usepackage{color}
\usepackage{lscape}
\usepackage{graphicx}
\usepackage[outdir=./]{epstopdf}
\usepackage{float}

\usepackage{graphicx}
\usepackage{caption}
\usepackage{subcaption}

\graphicspath{{./figures/}}

\usepackage{graphicx} 

\usepackage{amsmath}
\usepackage{amsthm}
\usepackage{amsfonts}
\usepackage{amssymb,amsmath}
\usepackage{epstopdf}
\usepackage[utf8]{inputenc}
\usepackage{algorithm}
\usepackage{algpseudocode}
\usepackage{multirow}
\usepackage{color}
\usepackage{algorithm,algpseudocode}

\algnewcommand{\algorithmicforeach}{\textbf{for each}}
\algdef{SE}[FOR]{ForEach}{EndForEach}[1]
  {\algorithmicforeach\ #1\ \algorithmicdo}
  {\algorithmicend\ \algorithmicforeach}

\usepackage[left=0.62in,right=0.62in,top=0.75in, bottom=1in]{geometry}

\begin{document}

\title{ A Comprehensive Survey On Client Selections in Federated Learning}
 
 \author{\IEEEauthorblockN{A. Gouissem$^1$, Z. Chkirbene$^2$, R. Hamila$^2$}

 \IEEEauthorblockA{$ ^1$   College of Computing and Information Technology, University of Doha for Science and Technology,  Qatar,}
\IEEEauthorblockA{$ ^2$ Electrical Engineering, Qatar University, Qatar,}
}
 
\IEEEoverridecommandlockouts
\IEEEpubid{\makebox[\columnwidth]{  979-8-3503-4760-9/23/\$31.00~\copyright{}2023 IEEE
\hfill} \hspace{\columnsep}\makebox[\columnwidth]{ }}
\maketitle
  \begin{abstract}
      Federated Learning (FL) is a rapidly growing field in machine learning that allows data to be trained across multiple decentralized devices. The selection of clients to participate in the training process is a critical factor for the performance of the overall system. In this survey, we provide a comprehensive overview of the state-of-the-art client selection techniques in FL, including their strengths and limitations, as well as the challenges and open issues that need to be addressed. We cover conventional selection techniques such as random selection where all or partial random of clients is used for the trained. We also cover performance-aware selections and as well as resource-aware selections for resource-constrained networks and heterogeneous networks. We also discuss the usage of client selection in model security enhancement. Lastly, we discuss open issues and challenges related to clients selection in dynamic constrained, and heterogeneous networks.
  \end{abstract}

  \section{Introduction}
In recent years, there has been a significant increase in the size and diversity of data sources available for analysis, including data generated by social media platforms, internet of things (IoT) devices, and a wide range of other sources. The growth in data size and sources created new challenges in terms of storage, processing, and analysis \cite{ li2022big}.

Classical analytical methods, which rely on statistical analysis and  and human-based decisions,  are nowadays limited in their ability to process and analyze data at scale, particularly when the data is complex or unstructured. The integration of machine learning (ML) techniques, on the other hand, can facilitate faster and more precise data analysis, resulting in better decision-making and increased efficiency. Additionally, it  can also help to uncover previously undetected patterns and trends that may not have been apparent using other methods. ML attract a lot of attention thanks to its ability to be applied in a wide range of areas, including image and speech recognition, relay selection \cite{gouissem2019machine}, natural language processing, intrusion detection \cite{chkirbene2020weighted, chkirbene2020iterative,chkirbene2020weighted2}, anti jamming, \cite{gouissem2022accelerated,gouissem2021game1,guizani2020combating} and even self-driving cars\cite{deshmukh2022overview}. 

While ML can facilitate advanced and efficient decision-making, it relies on the collection and centralization of data, which may not be possible due to privacy concerns \cite{zinamachine}. To address this challenge,  Federated learning (FL) is used to operate on decentralized datasets that are distributed across multiple devices. In particular, FL is a machine learning method that allows data to be trained across multiple decentralized devices, such as smartphones or laptops, without the need to centralize the data\cite{ banabilah2022federated}. 

FL is an emerging field in machine learning, with a growing body of literature exploring its potential benefits and challenges. Review papers such as \cite{ banabilah2022federated, lim2020federated, yang2019federated, zhang2022towards}, provide an overview of the concept and its applications, the state-of-the-art techniques in FL as well as the current challenges and open issues. However, while these papers have provided a good introduction to the field, there are still many challenges that need to be addressed. One of these challenges is the issue of client selection. In Federated Learning, the selection of clients to participate in the training process is critical to the performance of the overall system. The clients selected should be representative of the entire population and should have the necessary resources and capabilities to contribute to the training process. However, selecting the appropriate clients is a non-trivial task, particularly in large-scale networks with a large number of diverse clients.

Another challenge is the issue of privacy and security. Federated learning has the potential to improve privacy and security for the individuals or organizations contributing data. However, there are still many open questions regarding how to protect the privacy and security of the clients and their data in a federated learning setting. Additionally, there is a need for techniques that can detect and prevent malicious clients from joining the network, which can compromise the overall security of the system. {\color{black}Adaptivity} is also a key challenge in Federated Learning (FL) systems. In order to be effective, FL systems must be able to adapt to changing network conditions and client capabilities over time.  However, many current FL techniques lack this ability and are unable to adjust accordingly. This limits the effectiveness of the FL system, and it is an area of active research to improve the adaptivity of these systems.

This paper is a review of client selection techniques in FL. The goal of this paper is to provide a comprehensive overview of the state-of-the-art client selection techniques in FL, including their strengths and limitations, as well as the challenges and open issues that need to be addressed in this field. This review aims to provide a comprehensive understanding of the current state of client selection in FL and to provide insights for researchers and practitioners working in this field.

{\color{black}\section{Conventional  FL Selection }}


Several studies \cite{laguel2020device,mcmahan2017communication} have made substantial advancements in various aspects of FL, such as minimizing communication costs and addressing issues caused by non {\color {black} independent and identically distributed} (i.i.d) data distributions. However, most of these techniques rely solely on the selection of random clients which make the model performance non optimal. {\color {black} In this section, the  random selection as well as full and partial client selection will be thoroughly examined. The potential drawbacks of each method will also be discussed. }

\subsection{Random Selection}

{\color {black} Random selection in federated learning  refers to the process of randomly selecting a subset of devices or participants from a larger group to contribute data or computation to a FL model. This method of selection aims to provide a representative sample of data from the entire group and can be used to improve the generalization of the FL model.} The  authors in \cite{mcmahan2017communication} propose a method for training deep neural networks in a FL manner using iterative model averaging and conduct a thorough evaluation of its performance using different datasets. 
The proposed approach in \cite{mcmahan2017communication} considers  the challenges of client availability as well as  unbalanced and non-IID data. However, the client selection is performed randomly. In fact, the process begins with the server selecting a group of clients randomly. Each selected client then uses the global information and their own data to perform calculations locally, after which they send the update back to the server. The server takes in these updates and updates the global information, the cycle is then repeated with different clients. This increases the processing parallelism, where a larger number of clients can work independently between each round of communication. It also increases the computations per client in order to reduce the communication cost. The simulation results show that this approach is effective even when dealing with unbalanced and non-IID data and can reduce the number of communication rounds, by 10-100 times, as compared to other methods like synchronized stochastic gradient descent.

Consequently, most of the proposed schemes for FL models  rely on just selecting a random subset of participants from a large pool of available clients in each round of FL. This random selection strategy when combined with other algorithms and aggregation schemes can solve specific problems related to security, cost or data distribution.
{\color{black}The use of random selection in Federated Learning (FL) has its limitations. It does not take into account the variations in data, resources, and systems among clients, which can impact the efficiency of federated training. By randomly selecting clients, there is a risk of selecting clients with over-represented data or clients that are slow due to limited computation speeds or network bandwidths, which can act as bottlenecks in each round of federated training. This can lead to delays and inefficiencies in the training process \cite{cho2020bandit}.}
{\color{black}\subsection{Full and partial State Knowledge  Selection}}
{\color{black}In FL, full state knowledge means that all devices have complete knowledge of the global model, while partial state knowledge means some devices have limited knowledge and may not have access to the same information. In \cite{li2019convergence}, have demonstrated convergence in federated learning models when clients fully participate in the training process. However, these studies only apply to unbiased client participation. In \cite{ruan2021towards} evaluated the convergence of federated learning when clients have the ability to join or leave the training process or send incomplete updates to the server. Furthermore, most of the client selection techniques available in the literature assume the knowledge of the client-side state information\cite{abdelmoniem2021resource, cho2020bandit, cho2020client, goetz2019active, ji2020dynamic, lai2021oort, nishio2019client, ribero2020communication}. However, assuming full knowledge of the client-side state information in federated learning may not be feasible or practical to obtain such information in real-world scenarios. For example, in a large-scale federated learning system with thousands of clients, it may be difficult to collect and transmit detailed state information for each client. Additionally, obtaining such information may be a privacy concern for clients. This could lead to a low adoption rate, and less participation from clients.  }

\section{Performance Aware Selection}
{\color{black} In Performance aware selection is a technique that chooses a specific group of devices based on their expected performance during the training process, these devices are selected to improve the performance of the model.} Thus, the clients can be selected based on their statistical utility and their potential to improve the global model accuracy. Such decisions can be made based on the model updates using various measures \cite{cho2020client, goetz2019active, ribero2020communication, ouyang2021clusterfl}. {\color{black} In this section, various methods for selecting performance-aware federated learning techniques will be discussed.}
\subsection{Active Federated Learning}
{\color{black} Active Federated Learning  aims to improve the model's performance by selectively training on clients with valuable updates.} The authors in \cite{ goetz2019active} highlight that  FL training and update process consumes a large amount of the clients’ bandwidth which makes minimizing the transmission costs crucial to the success and efficiency of FL. Since the data on each client is highly variable, Active Federated Learning is proposed in \cite{ goetz2019active} to allow clients to be selected in each round based on the current model and client data to maximize efficiency, rather than being chosen uniformly at random. A cost-effective sampling method is also proposed to reduce the number of required training iterations by 20-70\% while maintaining the same model accuracy.
Active Federated Learning is a method that capitalizes on the stability and availability of most clients during the training process, by using client-side state information to guide the selection of clients. One such approach is ClusterFL \cite{ouyang2021clusterfl} that aims to improve the performance of the FL model by identifying and removing clients that are not performing as well as others or are less related to other clients within the same cluster. This approach utilizes the inherent clustering structure among clients based on their local data distributions to identify these clients and exclude them from the training process. However, Active Federated Learning can be computationally expensive to evaluate the model's performance on a large number of clients before deciding which clients to select for the next round of training

\subsection{UCB-CS}
{\color{black}Upper Confidence Bound for Cooperative Stochastic Bandits ( UCB-CS ) is a method for selecting clients in Federated Learning that balances exploration and exploitation using the UCB algorithm to select clients that are expected to provide the most valuable updates to the model. }In \cite{cho2020bandit} a new approach for selecting clients in FL that is based on the multi-armed bandit algorithm called UCB-CS is proposed.{\color{black}It addresses issues such as communication efficiency, noise, variations in local loss values, and error floors that are present in prior methods.} The authors show that  UCB-CS can achieve robustness to error floor, faster convergence, and reduced state estimates of local loss values, all without requiring additional communication compared to random selection. Additionally, it ensures fairness among clients by achieving more uniform performance. The work in \cite{cho2020bandit} is performed under the assumption of independent local losses for each client which might be unpractical in some scenarios since clients with similar characteristics may have similar local losses.  {\color{black} However, UCB-CS assumes that the rewards (or updates) provided by clients are independent and identically distributed, which may not be the case in some Federated Learning scenarios. }

\subsection{Contribution control mechanisms}
{\color{black}Contribution control mechanisms are methods for controlling the participation of clients to improve the training process.} The authors in \cite{ribero2020communication} propose an adaptive mechanism for selecting clients based on threshold recomputed at each round. By training the model locally and computing the local contribution of each device, the device itself can decide whether it should participate or not in the training according to a preset contribution threshold.

The work in \cite{ cho2020client} presents a study on the convergence of federated learning with a focus on biased client selection and its impact on training progress at each client. The authors found that bias towards clients with higher local losses improves convergence rates. They propose a new client selection strategy called Power-Of-Choice which results in faster convergence and better performance when compared to standard methods. 
The authors in \cite{li2022pyramidfl} proposed a technique named Pyramid that aims to accelerate the FL training process while maintaining a high model accuracy. PyramidFL performs client selection and prioritization while taking into account the dissimilarities between chosen and unselected clients. It also fully utilizes the data and system heterogeneity within the selected clients to optimize their contribution into the global model. To achieve this, PyramidFL first selects clients based on their utility from the global model and then improves the selection by optimizing client utility locally. The authors also report that PyramidFL was able to improve final model accuracy and speed at the same time. {\color{black} However, the model's performance in PyramidFL may also be limited by the quality of the clients in the lower levels of the pyramid, as these clients are likely to have less computational and communication resources.}

\section{Resource Aware Selection}
{\color{black} Resource-aware selection in Federated Learning is a technique that aims to select clients based on their available computational and communication resources}. The selection can also be performed while aiming to optimize the limited clients resources \cite{abdelmoniem2021resource, nishio2019client} and communication constraints \cite{ ji2020dynamic}. 

\subsection{Resource Constrained networks}

The authors in \cite{nishio2019client} proposed a federated learning (FL) protocol called FedCS, which is designed to mitigate the training bottleneck in mobile edge computing (MEC) frameworks. In the FedCS protocol, a server coordinates training in a cellular network comprising participating mobile devices with heterogeneous resources. The server gathers information about wireless channel states and computing capabilities from a subset of randomly selected participants, and then selects the maximum possible number of participants that can complete training within a prespecified deadline for the subsequent global aggregation phase.  A greedy algorithm is used in \cite{sviridenko2004note} to solve the maximization problem by iteratively selecting participants that take the least time for model upload and update. Simulation results showed that FedCS can achieve higher accuracy than an FL protocol that only accounts for training deadline, but FedCS has only been tested on simple deep neural network (DNN) models. There are limitations to the use of FedCS for more complex models, such as the difficulty in estimating the number of participants to be selected and potential bias towards selecting participants with devices that have better computing capabilities, which may not hold data representative of the population distribution.

\subsection{Reinforcement learning based selection}
One of the major challenges in implementing FL in mobile edge networks is that such networks are very dynamic and hard to predict \cite{chkirbene2021energy}. Therefore, static user selection techniques might fail to track the network and constraints variations. In particular, these participating mobile devices have limitations in terms of energy, processing power, and wireless bandwidth. The FL server must therefore determine the most efficient use of the resources to minimize energy consumption and training time. To solve this issue, the authors in \cite{anh2019efficient} proposed the use of Deep Q-Learning (DQL) to optimize the allocation of resources for training deep neural network models. The system includes mobile devices that collaborate to train the DNN models required by a FL server. These mobile devices are limited in terms of energy, processing power, and wireless bandwidth.  The Double Deep Q-Network (DDQN) algorithm can also be used as suggested in \cite{van2016deep}. The DDQN algorithm is applied to an optimization problem in which the FL server acts as the agent, the state space includes the energy and CPU state of the mobile devices, and the action space includes the amount of data and energy units that are used from the mobile devices. The rewards of this optimization problem is computed from factors such as data accumulation, energy consumption, and training latency. The results of the simulation studies show that this method can reduce energy consumption by about 31\% when compared with other commonly used methods such as the greedy algorithm and can decrease training latency by up to 55\% compared to random selection methods. 


Other research works, such as \cite{lim2020federated}, have focused on the difficulties of implementing Federated Learning in mobile edge networks and proposed potential solutions and use cases. For instance, \cite{park2019wireless} proposed a framework for effective resource management for FL in the edge. But, these studies, including \cite{park2019wireless,wang2019edge}, did not delve into the key challenges of creating incentives and optimizing network performance in edge-based FL.


\subsection{Heterogeneous Networks}

 The authors in \cite{lai2021oort} propose a technique named Oort that considers both data and system heterogeneity at the same time. Such approach creates a compromise between data usage and system efficiency, and it has been shown to have superior time-to-accuracy performance compared to random selection. However, this technique considers the differences between chosen and unselected clients, and it doesn't take into account the variations within the chosen clients.

Another approach, called hybrid-FL, has been proposed in \cite{yoshida2019hybrid} as a way to generate an approximately IID dataset using only a limited number of privacy-insensitive participants. This is done by selecting participants whose data can form an approximately IID dataset, then training a model on the collected IID dataset and merging it with the global model trained by all participants. The simulation results showed that this approach can improve classification accuracy even with a small percentage of participants sharing their data. However, it is important to note that this approach has the potential to compromise user privacy and security. To mitigate this, an incentive and reputation mechanism should be implemented to ensure that only trustworthy participants are allowed to upload their data.


\subsection{Game Theory and Incentive Mechanisms}
Incentive mechanisms are generally used to motivate a person or a device to take specific actions. Incentives can be either positive or negative. Positive incentives encourage actions by offering rewards, while negative incentives discourage bad behavior by imposing penalties.
The use of game theory to design incentive mechanisms has been extensively studied in other fields such as crowdsensing \cite{yang2012crowdsourcing}, edge computing \cite{liu2017incentive}, device to device communication \cite{li2015incentive}, opportunistic networks \cite{zhan2017incentive} and others, where users might be regraded for using or providing specific resources, and penalized for using others.  However, in FL, it is challenging to precisely quantify the value of each client's training data and to model the final learning performance of the system. As a result, it is difficult to model the utility function of each participant, which makes it hard to directly apply existing incentive mechanism design works to FL.
In contrast, \cite{khan2020federated} aims to comprehensively review resource optimization and incentive mechanisms for FL in edge networks.

The authors in \cite{khan2020federated} propose a novel approach to developing an incentive mechanism for Federated Learning (FL) using game theory by using the framework of a Stackelberg game in which FL users are the followers and the base station is the leader. In this game, FL users can strategically set the number of local iterations to maximize their own utility. The base station, as the leader, then uses the best response strategies of the users to optimize FL performance. This approach offers a new perspective on how to incentivize FL participants and improve overall FL performance. The paper also identifies key open research challenges and provides guidelines for FL in edge networks. {\color{black} However,  edge network architecture may be complex and distributed, which can make it difficult to optimize resources and implement incentive mechanisms. Moreover, FL based  on a Stackelberg game can raise privacy and security concerns, as data is exchanged between devices and organizations.}

 {\color {black}
\section{FL Client Selection Based Security}
 FL Client Selection Based Security ensures that only authorized clients and devices can access protected resources, and that sensitive data remains confidential. In this section, several techniques for selecting clients using security concepts in FL will be presented.}
\subsection{Poisoning attacks in FL}

Federated Learning (FL) is known to be vulnerable to poisoning attacks \cite{blanchard2017machine,guerraoui2018hidden,9771594}. The adversary’s goal and capabilities might differ from one model to another \cite{chkirbene2021data}. Based on this goal, poisoning attacks can be classified into targeted and untargeted attacks. In untargeted poisoning attacks, the objective is to lower the global model's accuracy on any test input \cite{guerraoui2018hidden}. In targeted poisoning, the objective is to reduce the accuracy of some test inputs while maintaining high accuracy for the remaining inputs. Therfore, untargeted assaults may drastically degrade the performance of the global model for any test input. Poisoning may be carried out either directly on the model (model poisoning) or by changing the training data. In untargeted poisoning attacks, the goal is to minimize the accuracy of the global model on any test input \cite{guerraoui2018hidden}. In targeted poisoning attacks, the goal is to minimize the accuracy on specific test inputs, while maintaining high accuracies on the rest of the test inputs. Untargeted attacks, however, can completely deteriorate the global model performance for any test input. The poisoning can be either performed on the model itself (model poisoning) attacks, \cite{guerraoui2018hidden}, or by manipulating the training data.

Several techniques have been proposed to combat such attacks. The main of focus of this survey in this regard is the investigation of the different techniques in the literature that perform the identification and isolation of the malicious nodes by selecting the most trustworthy ones data owners. In particular, several techniques exist in the literature to select the legitimate data owners. The different techniques in Section II and Section III perform the clients selection before the client training itself to avoid useless computations. However, poisoning attacks defense mechanisms perform the selection after the reception of the updates and by trying to identify anomalies in the reports.

\subsection{Krum}
{\color{black}Key-value Retrieval Under Misbehaviors (Krum)  aims to improve the robustness of FL in the presence of Byzantine or adversarial clients. Krum works by selecting a small number of clients to update the global model, and then using the median of their updates to update the global model.} The authors in \cite{blanchard2017machine} investigate a FL model when there are Byzantine workers present. The paper presents a Byzantine resilience client selection technique that allow the isolation of malicious nodes reports. In particular, one way to ensure this resilience is to consider a majority-based approach, but it has prohibitive computational costs. An alternative approach is the use of the KRUM metric that creates a combination of majority-based and squared-distance-based methods by choosing the vector that is closest to its $n-f$ neighbors. Krum has the advantage of having a local time complexity of $O(n^2 d)$ and is shown to converge in the presence of Byzantine workers.

\subsection{Multi-Krum}

An extension of the Krum method entitled Multi-Krum is also presented in \cite{blanchard2017machine}  and combines the resilience of Krum with the convergence speed of the standard averaging and performs well compared to other aggregation rules like the geometric median.
The major issue in both Krum and Multi-Krum is that the number of nodes to be isolated from the aggregation is setup in advance and fixed by the administrator whether there is a small or large number of malicious nodes in the network. Such approach might waste legitimate nodes time and resources if the number of malicious agents is smaller than the number of isolated clients. On the other hand, isolating few nodes results in an increased risk level.

\subsection{Bulyan}
Another method used to select benign clients and gradients is proposed in \cite{guerraoui2018hidden} and called Bulyan. The Bulyan algorithm insurance Byzantine-resilience aggregation by using recursive rules based on Krum or Brute, Krum, a Medoid, the geometric median or any other Byzantine–resilient rule. In fact, Among the set of all the users, the client with closest gradient (based on KRUM or any other rule) is first selected. This client is removed from the set and the process is repeated again till obtaining the right number of clients.
This approach is proven to be able to significantly reduce the space in which an attacker can cause the model to drift towards suboptimal models. Through experimental testing with different datasets, the authirs demonstrate that Bulyan successfully avoids convergence to ineffective models, instead resulting in models that are similar to those produced by a non-attacked averaging scheme

\subsection{Trimmed-mean}
Instead of selecting the same set of clients in all dimensions, the Trimmed-mean technique proposed in \cite{yin2018byzantine} and \cite{xie2018generalized} allows the selection of a different set of clients at each dimension. In particular, for each coordinate of the gradient, the  $\beta$ smallest and the $\beta$ highest reports are removed and not considered in the aggregation.
According to \cite{yin2018byzantine}, Trimmed-mean reaches an error rate that is close to optimal when m is approximately $m \leq \beta \leq \frac{n}{2}$ for objective functions that are strongly convex.
\subsection{Other techniques}
Several other techniques exist in the literature to protect the FL training from poisining attacks. For example, the authors in \cite{C} investigated a special type of data poisoning attack where Byzantine nodes may report high data sizes in order for the model aggregation to be unbalanced in their favor. A data balancing approach based on weight-truncation is proposed to protect the FL training. 
In \cite{A}, an enhanced version of the Alternating Direction Method of Multipliers (ADMM) algorithm is presented to ensure the convergence of decentralized learning algorithms to a suboptimal solution even when there is Byzantine data falsification. Additionally, the study includes an analysis of the speed of convergence of the algorithm.
The authors in \cite{D} analyze the dependency between iterations and aggregated gradients in order to identify and separate out Byzantine reported updates. A robust version of gradient descent is presented, which uses geometric median for model aggregation rather than traditional averaging. The mathematical analysis and simulation results in the paper support that this proposed method converges uniformly to the true gradient function.
However, all of the above techniques and several others in the literature process all the received reports to generate a Byzantine-free report without actually selecting or identifying the malicious ones. Such approaches might lack accuracy in some scenarios since the malicious nodes are nodes are not isolated and can still take part of the model direction decisions.

\section{Open issues and challenges}
{\color {black}In this section, we will discuss the open issues and challenges in FL to further improve the performance and effectiveness of Federated Learning systems.}
\subsection{Resource Constraints Challenges } One of the main challenges in FL is dealing with clients that have limited resources, such as computation power and network bandwidth. Random selection of clients may result in selecting clients that are not capable of contributing effectively to the training process due to their limited resources. In addition, there is  a limited number of studies that consider the dynamic nature of wireless networks. Numerous client selection strategies rely on static network circumstances, which may not correctly reflect the dynamic nature of wireless networks.  Furthermore, many client selection solutions do not account for the heterogeneous capabilities of the clients, which might result in reduced performances. In addition, numerous solutions rely on a centralized approach, which is sometimes impractical in large-scale networks due to the substantial computational and communication overhead. 

In general, most of the client selection approaches described in the literature for constrained networks are limited by their inability to adapt to dynamic network conditions, their disregard for client heterogeneous capabilities, and their centralized structure, which might be impractical for large-scale networks.
\subsection{Security Challenges}
Another challenge in FL is ensuring the security and privacy of the clients' data. Random selection of clients may result in selecting clients that are not trustworthy or that have vulnerable systems, which can compromise the security of the FL model. The available client selection techniques in the literature have several limitations. One limitation is that many techniques rely on a centralized approach, which can make the system vulnerable to attacks on the central entity.  Additionally, many techniques do not take into account the varying security capabilities of different clients, which can lead to suboptimal performance. For example, the trustworthiness of the clients should be different from one group to another based on their history and contribution to the global model. Furthermore, many techniques fail to consider the privacy concerns of the clients, which can lead to leakage of sensitive information. 

Another limitation is the lack of consideration for the dynamic nature of security threats and vulnerabilities, which can change over time. Additionally, many client selection techniques do not have the ability to detect or prevent malicious clients from joining the network in advance, which can compromise the overall security of the network.  In particular, most of the selection techniques in that regard let all the nodes train their models and then decide which ones to consider which might waster the client resources.

\subsection{Heterogeneity and Dynamicity Challenges}

In FL, clients can have different data distributions, system variations, and resource constraints. Random selection of clients may result in selecting clients that have over-represented data or clients that are not representative of the population, which can negatively impact the performance of the FL model. In particular, most of the current client selection techniques in literature have several limitations regarding heterogeneity and adaptivity. One limitation is that many techniques fail to take into account the dynamic nature of wireless networks and the dynamic capabilities of different clients, and rapidly changing network resources  which can lead to suboptimal performance and communication issues. In summary, Selecting clients dynamically based on the fast changing and heterogeneous state of the system and clients, the clients' resources and the model's performance is still an open challenge

\subsection{Performance trade-off Challenges}

Most of the client selection techniques in the literature focus on a single objective such as heterogeneity, security, or resources. {\color {black} The performance trade-off challenges in Federated Learning aim to balance between these objectives. } There was limited efforts in designing a multi-objective model that can handle several selection problem at once at the same time to create Performance trade-off by balancing the communication cost, model's performance and clients participation.

\section{Conclusion}
The selection of clients to participate in the training process is a critical factor for the performance of Federated Learning (FL) models. The clients selected should be representative of the entire population and have the necessary resources to contribute to the training process. However, choosing appropriate clients is a non-trivial task, particularly in large-scale networks with a large number of diverse clients. Poor client selection can lead to a decrease in the performance of the FL system, and increase in communication and computation cost. Furthermore, the clients selected should be trustworthy and reliable, as they may have malicious intent and attempt to poison the model by providing incorrect or biased data. In this survey, we have presented the state-of-the-art client selection techniques in FL, including their strengths and limitations, as well as the challenges and open issues. We covered conventional selection techniques, performance-aware selections, resource-aware selections, and model security techniques. We also discussed open issues and challenges related to resource constraints, heterogeneity and dynamicity, and performance trade-offs. We believe that addressing the open issues and challenges related to client selection will play a crucial role in the success and widespread adoption of FL in various applications.

\bibliographystyle{IEEEbib}
\bibliography{Mybiblio}

\end{document}